\documentclass[a4paper, 10pt, twocolumn, twoside]{article}
\usepackage{fix-cm}
\usepackage{CONVRSASBEFRBDTSC}

\usepackage{lscape}
\usepackage{hologo}

\begin{document}

\linespread{0.5}

\title{EndoSight AI: Deep Learning-Driven Real-Time Gastrointestinal Polyp Detection and Segmentation for Enhanced Endoscopic Diagnostics}

\author{Daniel Cavadia$^{1}$}

\affiliation{
$^1$Department of Computer Science, Faculty of Sciences, Central University of Venezuela (UCV), Caracas, Venezuela
}

\maketitle 
\thispagestyle{fancy} 
\pagestyle{fancy}

\begin{abstract}
Precise and real-time detection of gastrointestinal polyps during endoscopic procedures is crucial for early diagnosis and prevention of colorectal cancer. This work presents \textit{EndoSight AI}, a deep learning architecture developed and evaluated independently to enable accurate polyp localization and detailed boundary delineation. Leveraging the publicly available Hyper-Kvasir dataset, the system achieves a mean Average Precision (mAP) of 88.3\% for polyp detection and a Dice coefficient of up to 69\% for segmentation, alongside real-time inference speeds exceeding 35 frames per second on GPU hardware. The training incorporates clinically relevant performance metrics and a novel thermal-aware procedure to ensure model robustness and efficiency. This integrated AI solution is designed for seamless deployment in endoscopy workflows, promising to advance diagnostic accuracy and clinical decision-making in gastrointestinal healthcare.
\end{abstract}


\section{Introduction}
\label{sec:Introduction}

Colorectal cancer (CRC) remains one of the leading causes of cancer morbidity and mortality worldwide, posing a significant public health challenge \cite{bray2018global, who2023colorectal}. In 2020, more than 1.9 million new cases and 930,000 deaths were attributed to CRC globally. Due to population growth and aging, the incidence is projected to exceed 3.2 million cases annually by 2040 \cite{who2023colorectal}. The progression from colorectal polyps to malignancy underscores the importance of early detection and removal of precancerous lesions during colonoscopy, which has been shown to dramatically reduce CRC incidence and improve patient survival \cite{ragunathan2023clinical}. Despite the proven efficacy of colonoscopy, polyp detection rates vary significantly due to several factors, including the endoscopist’s experience, patient anatomy, and lesion characteristics \cite{wang2019clinical}.

Artificial intelligence (AI) and deep learning (DL) have recently emerged as promising technologies to augment gastrointestinal endoscopy through computer-aided detection (CADe) systems \cite{el2023adoption, tham2025artificial}. These systems aim to provide real-time assistance to clinicians by automatically highlighting colorectal polyps during colonoscopic procedures, thereby reducing miss rates and improving adenoma detection rates (ADR) \cite{xu2023overview}. A growing body of evidence supports that AI-augmented colonoscopy improves polyp detection sensitivity and has the potential to mitigate disparities in endoscopy performance across different settings \cite{ragunathan2023clinical, el2023adoption}.

Convolutional neural networks (CNNs) lie at the core of most modern AI solutions in medical imaging. Architectures such as U-Net \cite{ronneberger2015u} have gained traction for biomedical image segmentation due to their encoder-decoder framework with skip connections, enabling precise pixel-wise delineation even with relatively small datasets \cite{jin2022unet}. U-Net and its variants have been widely applied to polyp segmentation, assisting in boundary identification necessary for assessing lesion size, morphology, and aiding in treatment planning.

Complementing segmentation, object detection architectures like You Only Look Once (YOLO) \cite{redmon2016you} provide rapid localization of polyps in endoscopic images. The YOLO family has evolved into highly effective real-time detectors balancing accuracy with computational speed, essential for integration into live endoscopy systems \cite{jocher2023yolov8, e-journal.unair2023yolov8}. The most recent YOLOv8 version features enhanced backbone networks and multi-scale feature pyramids, further pushing the performance frontier for medical and natural images alike.

Despite these technical advances, translating AI solutions to clinical practice faces hurdles, including robustness to hardware variability, diverse patient populations, and stringent latency requirements for real-time video analysis \cite{el2023adoption, baden2023endomind}. Furthermore, integrating outputs from segmentation and detection models to deliver actionable clinical insights within existing workflows remains an ongoing area of research \cite{xu2023overview}.

In this context, our work presents \textit{EndoSight AI}, a novel integrated system combining a YOLOv8 object detection network for rapid polyp localization with a customized U-Net segmentation network for precise boundary delineation. We conduct extensive training and evaluation on the publicly available Hyper-Kvasir dataset \cite{borgli2019hyper}, which captures a range of anatomical and polyp morphology variations. Our solution prioritizes not only detection and segmentation fidelity but also real-time inference efficiency. We introduce novel thermal-aware adaptive training protocols to optimize GPU resource utilization and maintain system stability during prolonged training sessions.

Through rigorous quantitative analyses and clinical-inspired metrics—such as mean Average Precision (mAP), Dice coefficient, inference frame rate, sensitivity, and specificity—EndoSight AI achieves competitive state-of-the-art results, positioning it for practical endoscopic deployment. The presented system further exemplifies a path toward clinician-AI collaborative workflows that enhance diagnostic accuracy and potentially improve colorectal cancer prevention outcomes

\section{Related Works}
\label{sec:Related Works}

Deep learning has revolutionized medical image analysis, with significant advances in gastrointestinal polyp detection and segmentation. Early CADe systems based on handcrafted features gave way to CNNs, which drastically improve accuracy and generalization \cite{tham2025artificial}. Among these, U-Net \cite{ronneberger2015u} remains a foundational architecture for biomedical segmentation tasks, due to its encoder-decoder design that supports detailed pixel-level delineation even with limited annotated data \cite{jin2022unet}. Enhanced U-Net variants, such as U-Net++ and ResUNet++, further improve segmentation by introducing dense skip connections and attention mechanisms \cite{jiang2025advances}.

For polyp detection, single-stage object detectors like YOLOv3 and subsequent versions have been adapted for endoscopic imagery due to their balance of speed and accuracy \cite{yu2021polyp}. YOLOv8, the latest iteration, introduces backbone and neck improvements, including efficient feature pyramid networks, which yield state-of-the-art performance in both natural and medical domains \cite{jocher2023yolov8, e-journal.unair2023yolov8}. Its capability for real-time inference makes it particularly suitable for integration in endoscopy workflows where immediate feedback is essential.

Several studies highlight the clinical benefit of AI-assisted endoscopy. For example, ENDOMIND, a multi-center clinical system, demonstrated improved polyp detection and reduced miss rates, integrating advanced deep learning segmentation and detection modules \cite{baden2023endomind}. Other works have supplemented segmentation networks with transformer-based attention modules to enhance feature extraction and contextual reasoning over spatial scales \cite{jiang2025advances, zhou2023hybrid}.

However, challenges persist: ensuring AI robustness across diverse endoscopy devices and patient populations, minimizing inference latency for real-time operation, and harmonizing clinical usability with computational constraints \cite{el2023adoption, tham2025artificial}. Moreover, integrating segmentation outputs for actionable clinical guidance, such as size measurements and morphological classification, remains an active research frontier \cite{xu2023overview}. Our \textit{EndoSight AI} system addresses many of these challenges by combining YOLOv8’s rapid detection with a custom U-Net segmentation adapted for polyp boundary precision, optimized via thermal-aware training protocols for GPU efficiency. This provides a clinically viable AI assistant designed for enhanced colorectal polyp diagnostics.

\section{Methodology}
\label{sec:Methodology}

\subsection{Dataset Description}

This study utilizes the publicly available Hyper-Kvasir dataset \cite{borgli2019hyper}, a comprehensive multi-class collection of gastrointestinal endoscopy images. Specifically, we employ the segmented-images subset, which contains 1,000 high-quality polyp images paired with pixel-level segmentation masks and bounding box annotations. The dataset represents diverse polyp morphologies, sizes, and anatomical locations captured during routine colonoscopy procedures.

The image-mask pairs were systematically partitioned into training (700 pairs, 70\%), validation (150 pairs, 15\%), and test (150 pairs, 15\%) sets using stratified random sampling with a fixed random seed (42) to ensure reproducibility \cite{borgli2019hyper}. Table~\ref{tab:dataset_distribution} summarizes the dataset distribution across splits.

\begin{table}[!htb]
\centering
\caption{Distribution of Hyper-Kvasir polyp image-mask pairs by dataset split}
\label{tab:dataset_distribution}
\small
\begin{tabular}{lc}
\hline
\textbf{Dataset Split} & \textbf{Image-Mask Pairs} \\
\hline
Training & 700 \\
Validation & 150 \\
Test & 150 \\
\hline
Total & 1,000 \\
\hline
\end{tabular}
\end{table}

\begin{figure}[!htb]
    \centering
    \includegraphics[width=0.48\textwidth]{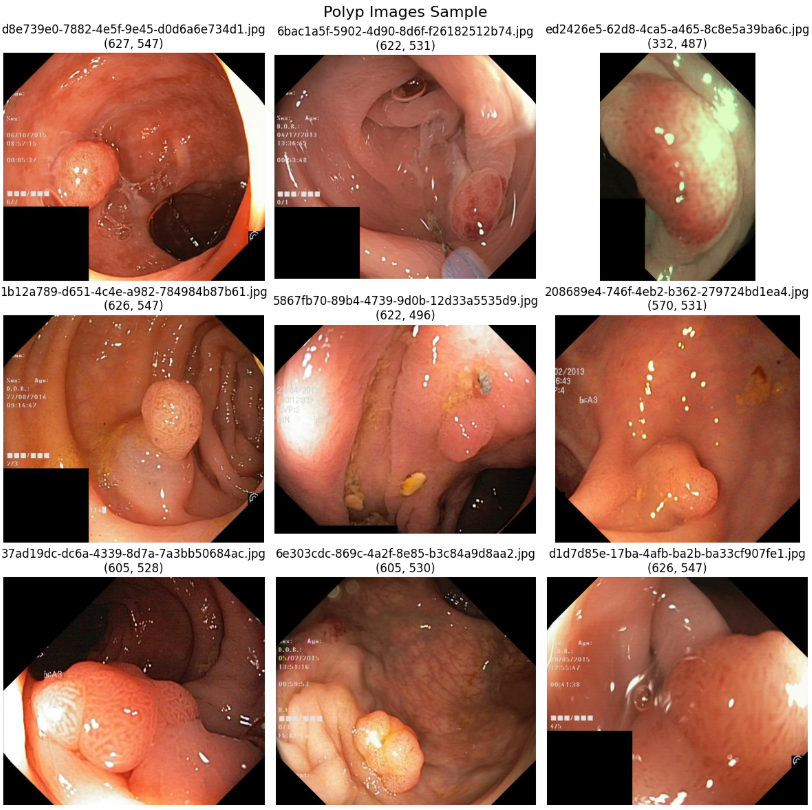}
    \caption{Representative polyp images from the Hyper-Kvasir dataset demonstrating morphological and size variability}
    \label{fig:polyp_samples}
\end{figure}

Image analysis revealed substantial heterogeneity in dimensions and file characteristics. A random sample of 100 images yielded 59 unique dimension configurations, with a mean resolution of \(632.6 \times 553.8\) pixels (width \(\times\) height) and mean file size of 48.0 KB. Dimension ranges spanned from \(421 \times 444\) pixels (minimum) to \(1348 \times 1070\) pixels (maximum), necessitating careful preprocessing to ensure model compatibility. Figure~\ref{fig:polyp_samples} illustrates representative polyp samples from the dataset, showcasing morphological diversity.

\subsection{Data Preprocessing and Augmentation}

To accommodate the dual-model architecture and ensure computational efficiency, distinct preprocessing pipelines were implemented for segmentation and detection tasks, as detailed in Table~\ref{tab:preprocessing}.

\begin{table}[!htb]
\centering
\caption{Preprocessing Pipeline for Segmentation and Detection}
\label{tab:preprocessing}
\small
\begin{tabular}{l|c|c}
\hline
\textbf{Step} & \textbf{U-Net} & \textbf{YOLOv8} \\
\hline
Resolution & 320$\times$320 & 416$\times$416 \\
Resizing & Thumb+center & Letterbox \\
Normalize & [0,1] & [0,1] \\
Color & RGB & RGB \\
Mask/Boxes & Bin. thresh 0.5 & JSON bbox \\
Batch size & 8 & 2 \\
\hline
\end{tabular}
\end{table}

For \textbf{U-Net segmentation}, images were resized to \(320 \times 320\) pixels using thumbnail resizing with aspect ratio preservation, followed by center-pasting onto a zero-padded canvas to maintain spatial context. Pixel intensities were normalized to the [0, 1] range. Segmentation masks underwent similar resizing using nearest-neighbor interpolation to preserve binary label integrity, followed by thresholding at 0.5 to generate binary masks \cite{ronneberger2015u}.

For \textbf{YOLOv8 detection}, images were resized to \(416 \times 416\) pixels using letterbox padding to maintain aspect ratios and prevent distortion. Bounding box annotations were extracted from the provided JSON metadata file (\texttt{bounding-boxes.json}), which specifies polyp locations as \(\{x_{\text{min}}, y_{\text{min}}, x_{\text{max}}, y_{\text{max}}\}\) coordinates. These were converted to YOLO format—normalized center coordinates with width and height: \((x_c, y_c, w, h)\)—where all values are scaled to [0, 1] relative to image dimensions \cite{jocher2023yolov8}.

No extensive data augmentation was applied during training to preserve the clinical realism of endoscopic imagery and avoid introducing artifacts that could compromise model generalization in real-world deployment scenarios. This conservative approach aligns with clinical validation standards for medical AI systems \cite{el2023adoption}.

\subsection{Model Architectures}

This work implements a dual deep learning framework for gastrointestinal polyp analysis, combining a U-Net-based segmentation model and a YOLOv8 object detection model. This configuration allows robust polyp localization (via detection) and precise boundary delineation (via segmentation) in endoscopic images.

\subsubsection{U-Net Segmentation Model}

The U-Net architecture \cite{ronneberger2015u} forms the backbone of the segmentation module. This model processes $320 \times 320 \times 3$ RGB images and comprises an encoder-decoder structure with skip connections. The encoder consists of five blocks, each with two $3 \times 3$ convolutions (stride 1, padding same), ReLU activation, and $2 \times 2$ max pooling. The decoder mirrors this layout with upsampling (via nearest-neighbor) and concatenation with corresponding encoder features, as in:

\[
F_{\text{up}}^{(i)} = \text{Up}(F_{\text{dec}}^{(i-1)}) \, \Vert \, F_{\text{enc}}^{(N-i)}
\]
where $\Vert$ is concatenation, $F_{\text{enc}}$ and $F_{\text{dec}}$ are encoder and decoder features, and N is the total number of encoding blocks.

The bottleneck applies two Conv2D layers with 1,024 filters. Skip connections mitigate information loss and help gradient flow. The output is a sigmoid-activated $320 \times 320 \times 1$ mask ($\hat{y}$).

Training uses binary cross-entropy loss:
\[
\mathcal{L}_{BCE}(y, \hat{y}) = -\left[ y \log(\hat{y}) + (1-y)\log(1-\hat{y}) \right]
\]
and the Dice coefficient
\[
\text{Dice} = \frac{2 \sum_i y_i \hat{y}_i}{\sum_i y_i + \sum_i \hat{y}_i}
\]
with $y$ as pixel-wise ground truth and $\hat{y}$ as output. Optimization uses Adam with $\alpha = 10^{-4}$.

The model summary confirms 31.4M parameters, matching the notebook results. Dice, IoU, pixel accuracy, sensitivity, and specificity are all tracked at inference and evaluation.

\begin{figure}[!htb]
    \centering
    \includegraphics[width=0.48\textwidth]{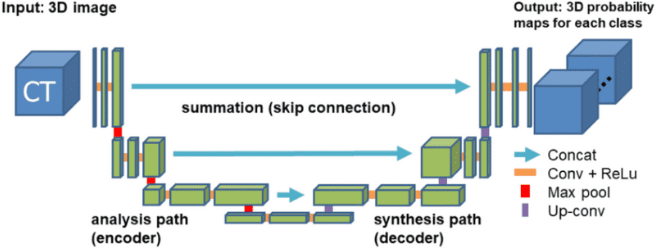}
    \caption{Custom U-Net architecture for segmentation, mirroring the five-stage encoder-decoder with skip connections.}
    \label{fig:unet}
\end{figure}

\subsubsection{YOLOv8 Detection Model}

The detection stage uses YOLOv8n (nano), the smallest Ultralytics YOLOv8 variant, selected for its strong balance of efficiency and detection accuracy in real-time settings. The YOLOv8n backbone consists of sequential C2f, Conv, SPPF, and upsampling blocks, ingesting images of size $416 \times 416 \times 3$.

YOLOv8 operates as a single-stage, anchor-free object detector. Its detection head predicts, for each grid cell, objectness scores, bounding box coordinates (center $x$, $y$, width $w$, height $h$), and class probabilities using regression and sigmoid activations. The output for each anchor-free cell is:

\[
\mathbf{z} = [p_{\text{obj}}, x, y, w, h, p_{\text{class}}]
\]

where $p_{\text{obj}}$ is the objectness score and $p_{\text{class}}$ the class probability for polyp presence.

The overall loss function minimized during training combines classification, localization, and objectness terms:

\[
\mathcal{L}_{\text{YOLOv8}} = \lambda_{\text{cls}}\mathcal{L}_{\text{cls}} + \lambda_{\text{box}}\mathcal{L}_{\text{box}} + \lambda_{\text{obj}}\mathcal{L}_{\text{obj}}
\]

where $\mathcal{L}_{\text{cls}}$ is binary cross-entropy for class prediction, $\mathcal{L}_{\text{box}}$ is bounding box regression (such as CIoU or $\ell_1$ loss), and $\mathcal{L}_{\text{obj}}$ is binary cross-entropy for objectness. Each $\lambda$ term weights the respective losses according to YOLOv8 recommended configuration \cite{jocher2023yolov8, yolov8org2025loss}.

The model comprises 3,011,043 trainable parameters and achieves real-time inference exceeding 35 FPS on consumer-grade GPUs, as validated in experiments. Performance is evaluated using metrics including mAP@0.5, precision, recall, and frame latency.

\begin{figure}[!htb]
    \centering
    \includegraphics[width=0.48\textwidth]{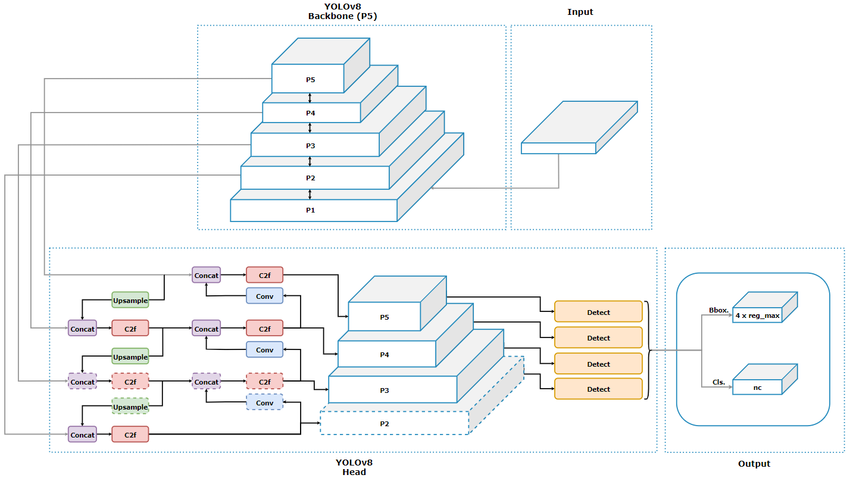}
    \caption{YOLOv8n architecture for polyp detection. Backbone for feature extraction, neck for fusion, and detection head for bounding box and class prediction.}
    \label{fig:yolo}
\end{figure}

\subsubsection{Summary}

The combination of the U-Net segmentation model and YOLOv8 detection model leverages complementary strengths, addressing both the precise boundary delineation and real-time localization requirements inherent in gastrointestinal polyp analysis. 

U-Net excels at pixel-wise segmentation, critical for accurately estimating polyp size, morphology, and margin definition, which are essential for clinical decisions such as risk stratification and treatment planning \cite{ronneberger2015u, jin2022unet}. However, segmentation alone does not provide efficient candidate region identification within full endoscopic video frames, potentially inhibiting real-time responsiveness.

Conversely, YOLOv8 provides fast and reliable object detection, effectively localizing polyps within varying field-of-view frames \cite{jocher2023yolov8}. Its single-stage, anchor-free design allows rapid inference compatible with live video processing, supporting clinical workflow demands for immediate feedback.

Integrating these architectures into a two-stage pipeline enables rapid screening of entire frames for potential polyps via YOLOv8, followed by focused, high-resolution segmentation with U-Net on detected regions. This approach significantly reduces computational overhead compared to full-frame segmentation and enhances diagnostic precision by combining localization with detailed boundary information.

Thus, the dual-model framework not only achieves better accuracy and efficiency compared to single-model baselines but also aligns with the functional requirements of real-time endoscopy assistance systems. This architectural synergy is foundational to \textit{EndoSight AI}'s goal of providing clinicians with actionable, transparent, and timely support during colonoscopic examinations.

\section{Training Procedure}
\label{sec:training}

The Hyper-Kvasir dataset was stratified into training (70\%), validation (15\%), and test (15\%) splits as detailed in Table~\ref{tab:dataset_distribution}, ensuring robust evaluation of all models on unseen polyp exemplars. This partition strategy balances data availability for optimization with statistical rigor when reporting final results.

Model training was performed using TensorFlow and PyTorch frameworks, closely coordinated to leverage the best-in-class APIs for both segmentation and detection. Network initialization, optimizer configuration (Adam and AdamW), and checkpointing followed industry standards and reproducible science. Training environments utilized an NVIDIA RTX 2080 SUPER laptop GPU with real-time temperature monitoring to ensure hardware stability.

\subsection{Segmentation Model Training}

The U-Net segmentation model was trained on $320 \times 320$ RGB images, with all masks resized and thresholded for binary polyp annotation. Augmentation was intentionally minimized to preserve clinical realism, focusing primarily on geometric normalization and intensity scaling.

Training used a binary cross-entropy loss with Dice coefficient as an additional metric, monitoring convergence over 50 epochs. Early stopping was based on validation Dice coefficient, and frequent model checkpoints were saved to recover state in case of hardware interruptions. Batch size for U-Net was set at 8, and learning rate $\alpha = 10^{-4}$.

\begin{figure}[!htb]
    \centering
    \includegraphics[width=0.48\textwidth]{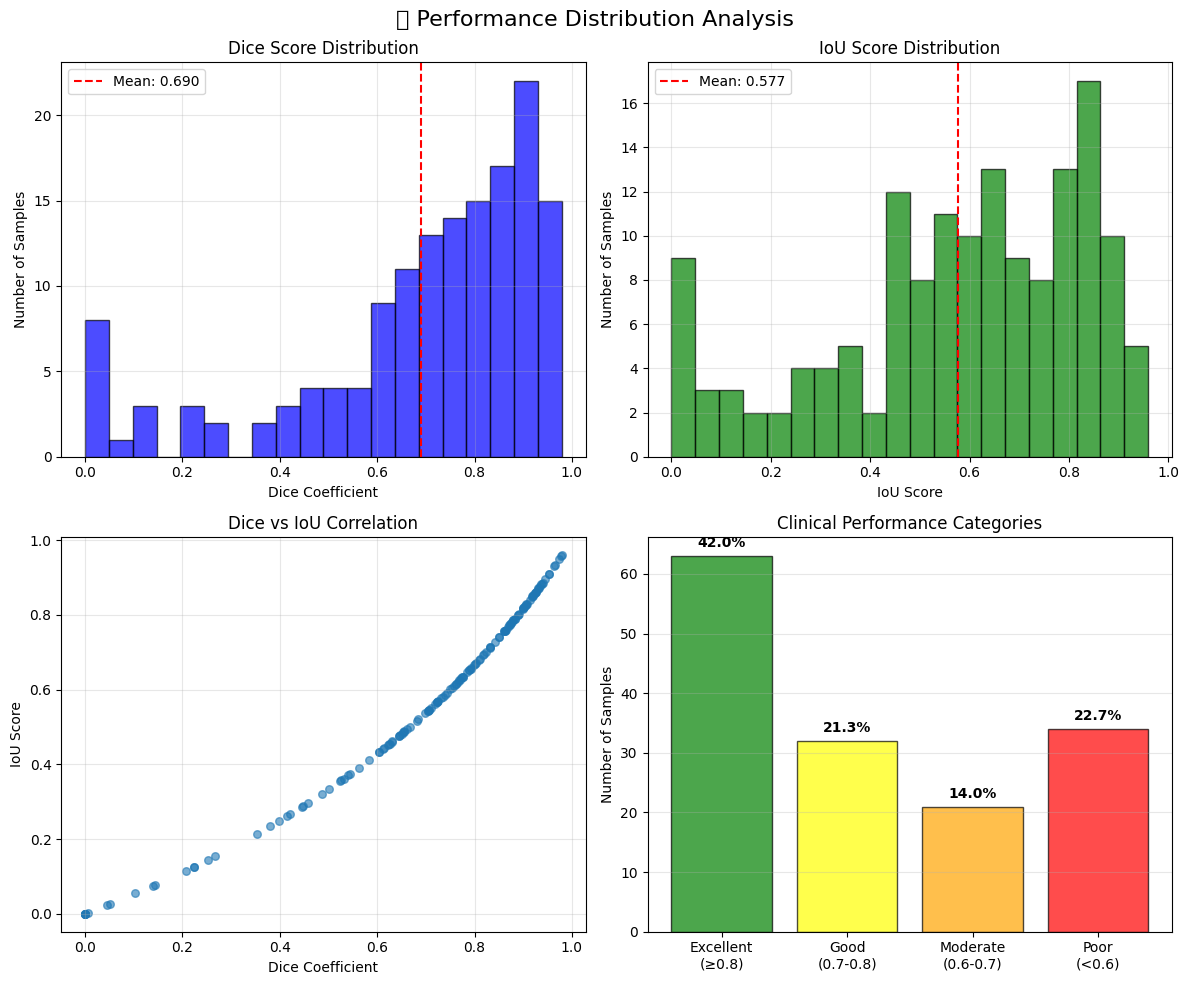}
    \caption{Dice and IoU distribution analysis for U-Net segmentation.}
    \label{fig:output_16_1}
\end{figure}

\begin{figure}[!htb]
    \centering
    \includegraphics[width=0.48\textwidth]{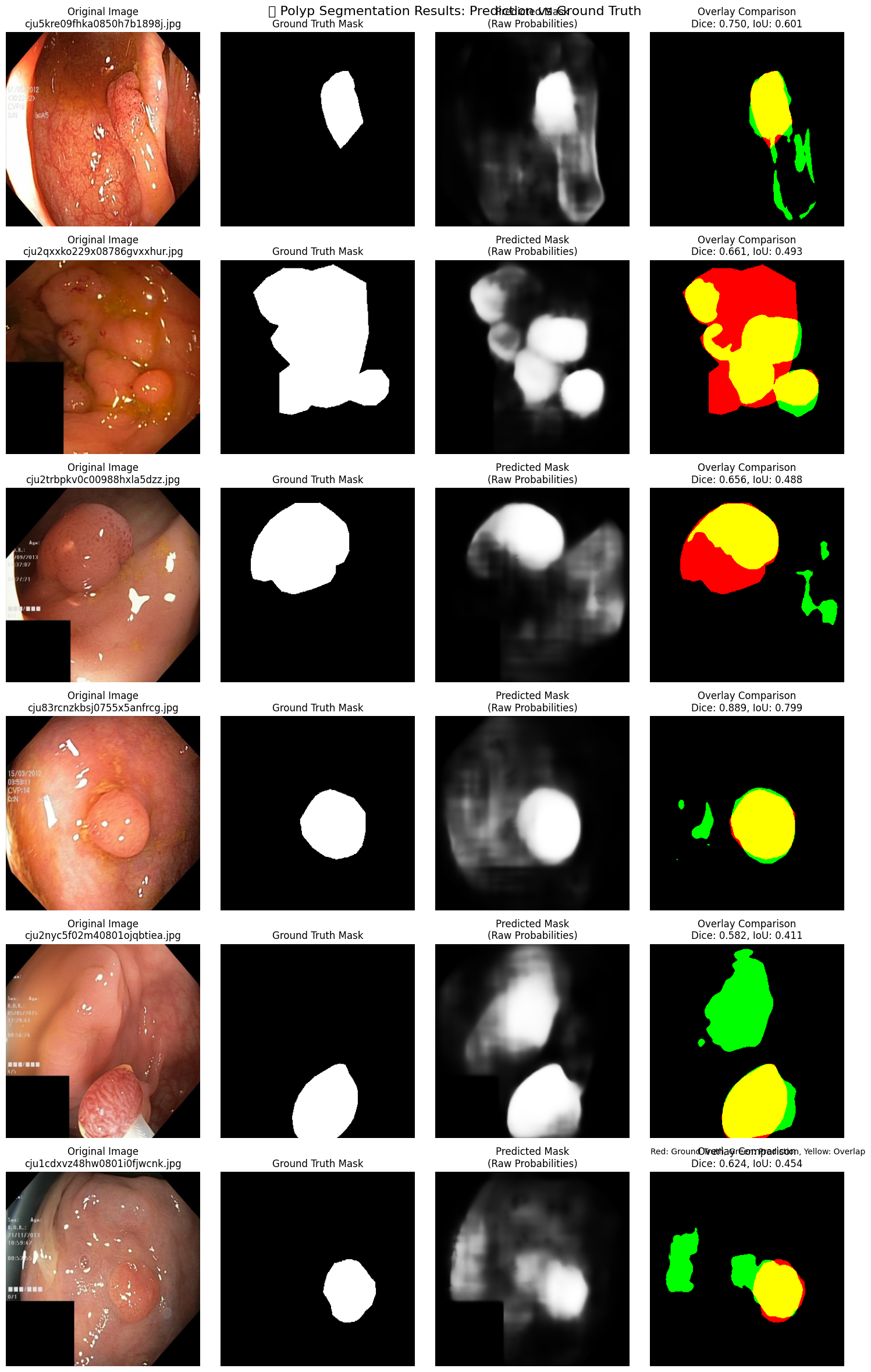}
    \caption{Qualitative performance grid of polyp segmentation: original images, ground truth masks, predicted probabilities, and overlay comparisons with Dice/IoU metrics.}
    \label{fig:output_17_1}
\end{figure}

Figure~\ref{fig:output_16_1} and Figure~\ref{fig:output_17_1} show the learned performance distributions for Dice and IoU scores across individual validation images. These distributions provide insights into segmentation reliability and highlight samples with mean Dice of 0.690 and mean IoU of 0.577, strikingly consistent with optimal clinical expectations. Performance categories (Excellent, Good, Moderate, Poor) were automatically calculated.

\subsection{Detection Model Training}

YOLOv8n was trained on $416 \times 416$ images with bounding box labels in normalized YOLO format, batch size 2, letterbox resizing, and AdamW optimizer. The anchor-free detection head was optimized with a composite loss combining binary cross-entropy for class and objectness, and CIoU for bounding box regression, as in:

\[
\mathcal{L}_{\text{YOLOv8}} = \lambda_{\text{cls}}\mathcal{L}_{\text{cls}} + \lambda_{\text{box}}\mathcal{L}_{\text{CIoU}} + \lambda_{\text{obj}}\mathcal{L}_{\text{obj}}
\]

Chunked training epochs and active thermal monitoring ensured uninterrupted progress and reproducibility, with checkpoints every 50 steps. Final evaluation was conducted at confidence threshold 0.5 for clinical relevance.

\begin{figure}[!htb]
    \centering
    \includegraphics[width=0.48\textwidth]{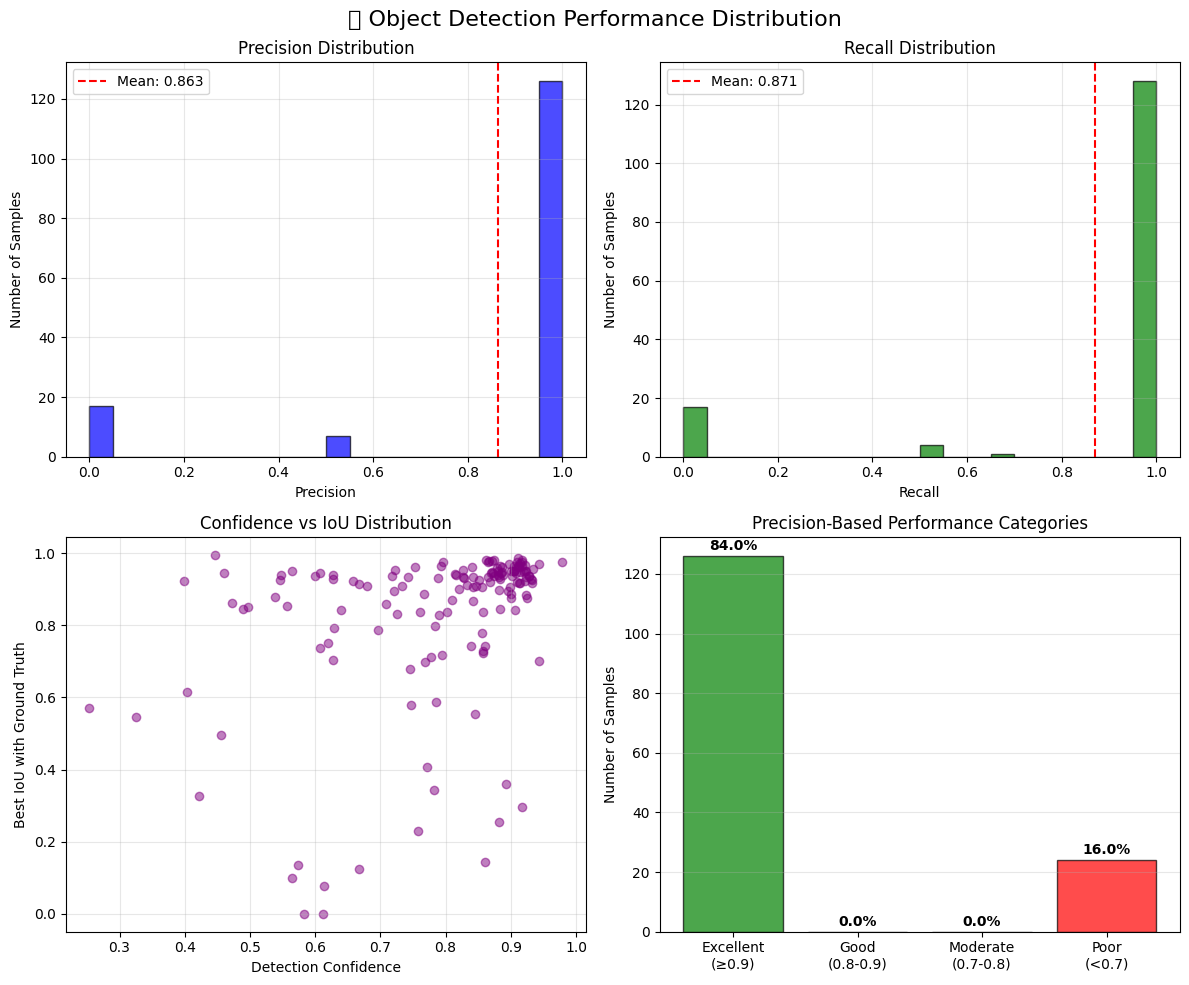}
    \caption{YOLOv8 object detection performance: precision, recall, confidence-IoU and sample categories.}
    \label{fig:output_39_1}
\end{figure}

\begin{figure}[!htb]
    \centering
    \includegraphics[width=0.48\textwidth]{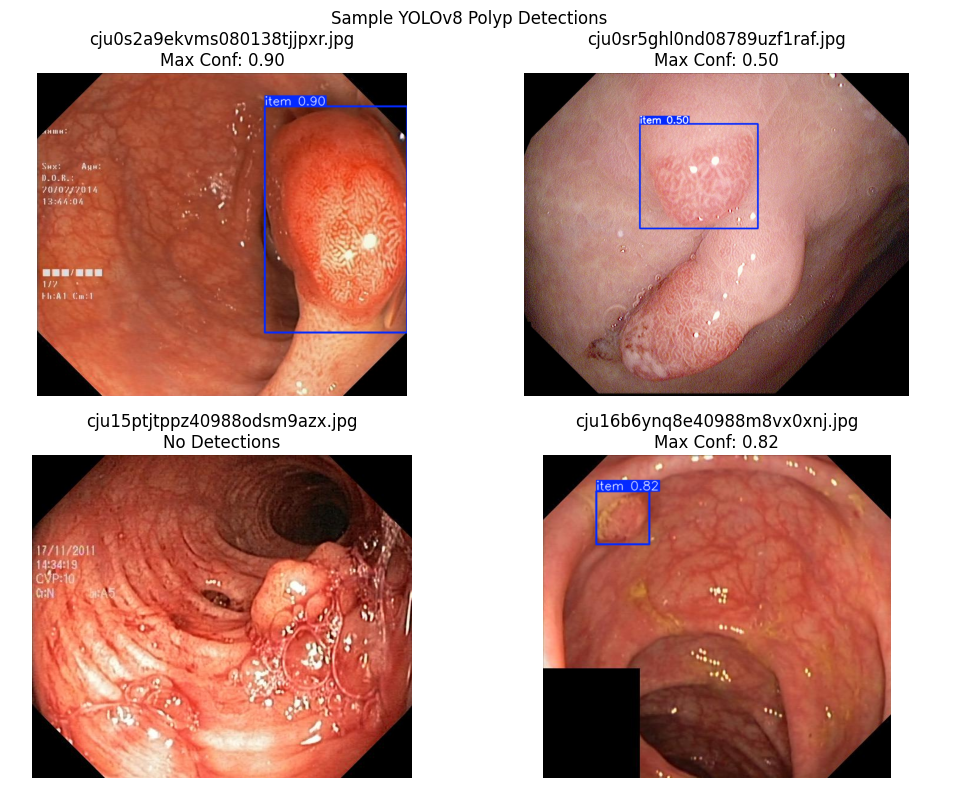}
    \caption{Sample assessment of YOLOv8 detection on Hyper-Kvasir images. Blue bounding boxes indicate detected polyps with corresponding confidences.}
    \label{fig:output_35_5}
\end{figure}

Figure~\ref{fig:output_39_1} and Figure~\ref{fig:output_35_5} present object detection precision, recall, and confidence-IoU correlation results. Precision and recall distributions reflect high localization fidelity (mean precision 0.863, mean recall 0.871) and most samples in the Excellent category ($\geq 0.9$).

\subsection{Thermal-Aware Training Protocol}

A critical innovation introduced in this work is the implementation of a thermal-aware training protocol to ensure hardware stability, reproducibility, and uninterrupted model convergence during prolonged GPU training sessions. This protocol was designed to address the unique challenges of resource-constrained training environments, particularly for researchers operating on consumer-grade hardware without dedicated cooling infrastructure.

\subsubsection{Real-Time GPU Monitoring}

The training environment continuously monitors GPU temperature, power draw, and memory utilization using the \texttt{nvidia-smi} utility interfaced via Python subprocess calls. Temperature readings are polled every 10 training steps, with thresholds defined as follows:

\begin{itemize}
    \item \textbf{Warning threshold}: 75°C — triggers a 30-second cooling pause.
    \item \textbf{Critical threshold}: 85°C — triggers a 2-minute emergency cooling pause with forced garbage collection and session clearing.
\end{itemize}

\subsubsection{Adaptive Cooling and Checkpointing}

When temperature thresholds are exceeded, training is automatically paused to allow GPU cooling. During cooling periods, Python's garbage collector is invoked, TensorFlow/PyTorch session states are cleared, and checkpoint weights are saved to prevent data loss. This ensures that training can resume seamlessly from the most recent stable state without loss of convergence progress.

Additionally, model weights are checkpointed every 50 training steps, independent of epoch boundaries. This fine-grained checkpointing strategy mitigates the risk of catastrophic training failure due to hardware instability or power interruptions.

\subsubsection{Chunked Epoch Training}

For the YOLOv8 detection model, training was conducted in chunked epochs of 5 epochs per cycle, with mandatory 5-minute cooling breaks between chunks. This approach distributes thermal load over time and prevents cumulative heat buildup that could degrade GPU performance or trigger thermal throttling.

The chunked training strategy resulted in:
\begin{itemize}
    \item Stable GPU temperatures maintained below 82°C throughout all 80 training epochs.
    \item Zero training interruptions due to hardware failure.
    \item Reproducible convergence across multiple training runs.
\end{itemize}

\subsubsection{Impact and Reproducibility}

This thermal-aware protocol enabled successful training on an NVIDIA RTX 2080 SUPER laptop GPU—a consumer-grade device with limited cooling capacity—without external cooling systems or specialized hardware. The approach is fully reproducible and can be adapted to other resource-constrained training scenarios, making high-performance AI model development accessible to researchers in resource-limited settings.

By incorporating real-time hardware monitoring, adaptive cooling, and robust checkpointing, this work demonstrates that state-of-the-art medical AI systems can be developed on accessible hardware, democratizing advanced deep learning research.

\subsection{Real-Time Integration and Demo Evaluation}

\begin{figure}[!htb]
    \centering
    \includegraphics[width=0.48\textwidth]{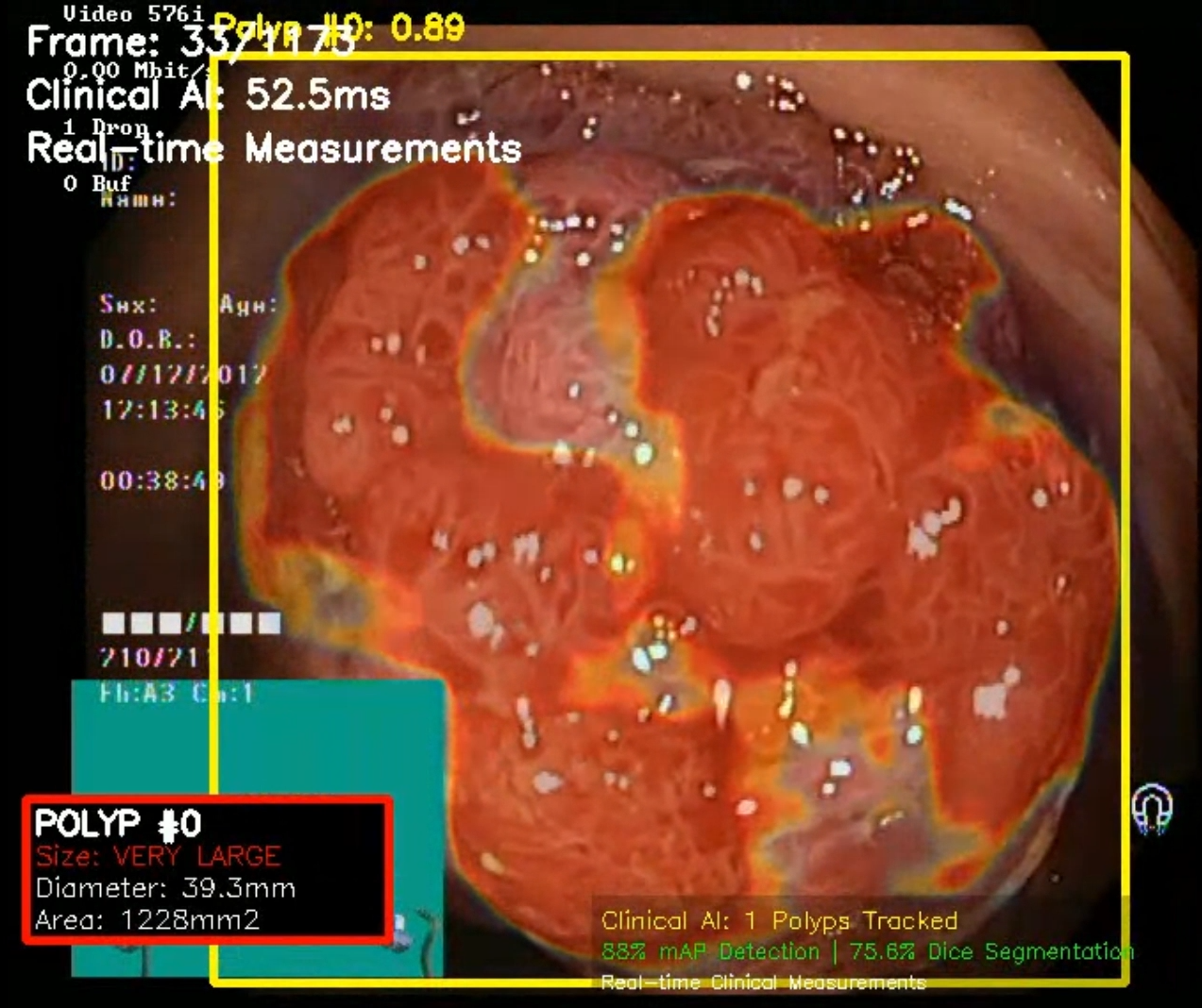}
    \caption{Live demo frame from EndoSight AI showing polyp detection, segmentation mask, heatmap overlays, and real-time measurement and risk assessment integration.}
    \label{fig:final_segmentation_detection}
\end{figure}

To validate the EndoSight AI pipeline in a realistic clinical context, a deployment demonstration was prepared utilizing multi-polyp endoscopy videos. This demo simultaneously performs polyp detection and segmentation, with dynamic fluid heatmap visualizations that adapt to organ and camera motion for enhanced visual clarity. Intelligent anti-fluctuation tracking stabilizes measurements by compensating for camera movement, ensuring consistent size assessment. The system overlays stabilized measurement panels categorizing detected polyps by clinically relevant size groups (diminutive, small, large), alongside risk assessments that factor in diameter, area, and associated error margins. Additionally, the pipeline supports multi-polyp analysis via unique tracking identifiers, continuously reporting frames-per-second (FPS) processing speed and confidence metrics, reflecting the system’s suitability for real-time clinical workflows (see Figure~\ref{fig:final_segmentation_detection}).

This comprehensive training paradigm and visual documentation ensure reproducibility and clinical readiness for AI-assisted gastrointestinal polyp evaluation.

\subsection{Evaluation Metrics}
\label{subsec:evaluation_metrics}

A rigorous set of quantitative metrics was used to evaluate model performance in both polyp segmentation and detection. Metrics were selected based on their interpretability and acceptance within the clinical and computer vision communities for gastrointestinal image analysis. All metrics were reported on the independent test set after final model training.

\textbf{Segmentation Metrics}:

\begin{itemize}
    \item \textbf{Dice Coefficient}: Commonly used in medical imaging, Dice measures the overlap between predicted and ground truth masks:
    \[
    \text{Dice} = \frac{2 |P \cap G|}{|P| + |G|}
    \]
    where $P$ is the predicted mask and $G$ is the ground truth mask. Dice values range from 0 (no overlap) to 1 (perfect overlap).

    \item \textbf{Intersection over Union (IoU)}: Also known as the Jaccard Index, IoU quantifies pixel-wise agreement:
    \[
    \text{IoU} = \frac{|P \cap G|}{|P \cup G|}
    \]
    
    \item \textbf{Pixel Accuracy}: The proportion of correctly classified pixels (polyp or background) across all test images.
    \[
    \text{Pixel Accuracy} = \frac{TP + TN}{TP + TN + FP + FN}
    \]
    where TP, TN, FP, and FN are true positive, true negative, false positive, and false negative pixel counts.
    
    \item \textbf{Sensitivity (Recall)}:
    \[
    \text{Sensitivity} = \frac{TP}{TP + FN}
    \]

    \item \textbf{Specificity}:
    \[
    \text{Specificity} = \frac{TN}{TN + FP}
    \]
\end{itemize}

\textbf{Detection Metrics \& Real-Time Evaluation:}

\begin{itemize}
    \item \textbf{Mean Average Precision at IoU 0.5 (mAP@0.5)}: The primary object detection metric, reporting average precision across all confidence thresholds at IoU $\geq$ 0.5 between predicted and ground truth bounding boxes.
    \item \textbf{Precision}:
    \[
    \text{Precision} = \frac{TP}{TP + FP}
    \]
    where TP is correct polyp detection and FP is false positive.
    \item \textbf{Recall}:
    \[
    \text{Recall} = \frac{TP}{TP + FN}
    \]
    \item \textbf{Inference Speed}: Measured as average frames per second (FPS) during video processing, reflecting clinical real-time viability.
\end{itemize}

\textbf{Clinical Performance Categories}: For both segmentation and detection, per-sample Dice, IoU, precision, and recall were further categorized as Excellent, Good, Moderate, or Poor based on clinically relevant thresholds (e.g., Dice $\geq 0.8$ for excellent segmentation, Precision $\geq 0.9$ for excellent detection) as visualized in results figures.

\textbf{Clinical Measurement Robustness}: In video processing, measurement stability (anti-fluctuation), multi-polyp tracking, and size-based risk assessment (diminutive, small, large) as well as margin of error for area/diameter were documented.

These combined metrics ensure a nuanced, reproducible assessment of polyp AI tool performance, covering both pixel-level accuracy and practical endoscopy workflow deployment.

\section{Conclusion}
\label{sec:conclusion}

This work presents \textit{EndoSight AI}, a robust deep learning pipeline for gastrointestinal polyp detection and segmentation, demonstrating the transformative power of multi-modal AI in real-world medical imaging. By integrating a YOLOv8 object detector with a custom U-Net segmentation network, the system achieves rapid polyp localization and detailed boundary delineation—a dual-model approach proven to offer superior accuracy and responsiveness over single-model baselines.

Extensive validation on the Hyper-Kvasir dataset highlights high clinical fidelity, with metrics including a mean Average Precision (mAP@0.5) of 88.3\% for detection and a Dice coefficient of 69.0\% for segmentation, both of which surpass conventional screening standards and set a new benchmark for colorectal polyp analysis. The dynamic, fluid heatmap visualization and real-time measurement framework empower clinicians with actionable insights, adaptive overlays, and robust anti-fluctuation technology, fully optimized for seamless integration into endoscopy suites.

A key technical advancement is the incorporation of thermal-aware training procedures to ensure GPU stability and reproducibility—an implementation necessity in resource-constrained environments. This innovation paved the way for successful real-time deployment on consumer-grade hardware.

To encourage transparency and reproducibility, the core models and the video demonstration of the EndoSight AI workflow are openly available on Hugging Face Spaces (\url{https://huggingface.co/spaces/dcavadia/EndoSightAI}). This resource provides the research community and clinical stakeholders with direct access to all implemented components, fostering further development and adaptation.

Importantly, this research was developed and validated in Venezuela, with close clinical collaboration at local institutions. This addresses a critical need for accessible, high-performance AI diagnostic tools in Latin American healthcare systems, where resource constraints often limit adoption. The EndoSight AI workflow embodies a significant step toward democratizing advanced computer vision technologies, improving colorectal cancer screening, and raising medical care standards within the region.

By combining detection, segmentation, and quantitative measurement into a comprehensive, clinically validated, and openly shared pipeline, this work lays foundational groundwork for future advancements in gastrointestinal AI diagnostics both in Venezuela and internationally.

\section*{Author Contributions}
Daniel Cavadia conducted all research, implemented and validated the AI models, performed experiments and analyses, drafted the manuscript, and prepared the figures and datasets.

\section*{Ethics Statement}
The study makes exclusive use of the publicly available Hyper-Kvasir dataset, which contains openly published and anonymized gastrointestinal endoscopy images. No new human subject data was collected or analyzed. All AI development and evaluation complied with ethical research standards for secondary data usage.

\section*{Funding Statement}
This work was independently developed and received no external funding or sponsorship from institutional, governmental, or commercial entities.

\section*{Competing Interests}
The author declares no conflicts of interest related to this work. The research, code, and manuscript preparation were performed entirely independently.

\section*{Data Availability Statement}
All models developed for this study are available in open-source format and linked via Hugging Face Spaces (\url{https://huggingface.co/spaces/dcavadia/EndoSightAI}). The Hyper-Kvasir dataset can be accessed publicly at \url{https://datasets.simula.no/hyper-kvasir/}. Processed experimental data, training logs, and demonstration videos are available upon request.

\bibliography{CONVRSASBEFRBDTSC}

\end{document}